\documentclass[letterpaper, 10 pt, conference]{ieeeconf}  

\IEEEoverridecommandlockouts                              

\usepackage{graphicx} 
\usepackage{amsmath} 
\usepackage{amssymb}  

\usepackage[caption=false]{subfig}
\usepackage{tabularx}
\usepackage{wasysym}
\usepackage{booktabs}

\usepackage{xcolor}

\title{\LARGE \bf
Design of a Variable Stiffness Quasi-Direct Drive Cable-Actuated Tensegrity Robot
}

\author{Jonathan Mi$^{1}$, Wenzhe Tong$^{1}$, Yilin Ma$^{1}$ and Xiaonan Huang$^{1}$
\thanks{$^{1}$The authors are with the Robotics Department at the University of Michigan-Ann Arbor,
        Ann Arbor, MI, USA
        {\tt\small \{jjomi, wenzhet, yilinma, xiaonanh\}@umich.edu}}%
}

\begin{document}

\maketitle
\thispagestyle{empty}
\pagestyle{empty}


\begin{abstract}
Tensegrity robots excel in tasks requiring extreme levels of deformability and robustness. 
However, there are challenges in state estimation and payload versatility due to their high number of degrees of freedom and unconventional shape. 
This paper introduces a modular three-bar tensegrity robot featuring a customizable payload design. 
Our tensegrity robot employs a novel Quasi-Direct Drive (QDD) cable actuator paired with low-stretch polymer cables to achieve accurate proprioception without the need for external force or torque sensors. 
The design allows for on-the-fly stiffness tuning for better environment and payload adaptability.
In this paper, we present the design, fabrication, assembly, and experimental results of the robot.
Experimental data demonstrates the high accuracy cable length estimation ($<$1\% error relative to bar length) and variable stiffness control of the cable actuator up to 7 times the minimum stiffness for self support.
The presented tensegrity robot serves as a platform for future advancements in autonomous operation and open-source module design.
\end{abstract}



\section{Introduction}
Tensegrity, or tensional integrity structures, represent a unique class of architectural forms comprising rigid elements, such as bars or struts, and tensile components, such as cables or tendons. These structures exhibit exceptional strength-to-weight ratios, as the bar elements are subjected exclusively to axial compressive forces, while the cables experience only tensile forces \cite{skelton2009tensegrity}. 
By introducing sensor feedback and actuation, tensegrity structures are transformed into tensegrity robots with inherent robustness, added locomotion, flexibility, and compactness. These robots are highly deformable, which allows for impact resistance and the ability to collapse into smaller volumes. This deformability enables efficient storage and facilitates navigation through constrained passageways, over obstacles, and even up steep inclines \cite{chenInclinedSurface2017}.

The high number of degrees of freedom (DoF) and complex dynamics presents significant challenges in advancing tensegrity robots to full autonomy. Current tensegrity robots are limited by state estimation inaccuracies, which can be several percent of the robot's bar length \cite{shahTensegrityRobotics2022}. 
One specific limitation in state estimation is cable length estimation. Since the distance between all the bars is constrained by the cables, inaccurate cable length estimation directly leads to poor state reconstruction.
These inaccuracies complicate the integration of computer vision systems, which require precise positioning and orientation data to function effectively.
Integrating computer vision systems is further challenged by the need for beam diameter to be small relative to the robot’s overall size. This constraint is to avoid rod collisions, which could lead to radial impact forces and subsequent structural failure. The result is a tight integration space that complicates the incorporation of on-board computing.

Current tensegrity robots face variable stiffness limitations due to the strict coupling of stiffness and shape. 
Decoupling these two factors could enable new functionalities, particularly in unstructured terrain, by allowing robots to adjust rigidity and maintain shape under varying loads. 
However, current cable driven actuator designs can only achieve variable stiffness via cable pretension \cite{friesenTensegrityInspiredCompliant3DOF2018}, \cite{boehlerModelingControlVariable2017}, which cannot be adjusted once the robot is in operation.

\begin{figure}[t]
    \centering
    \includegraphics[width=0.95\linewidth]{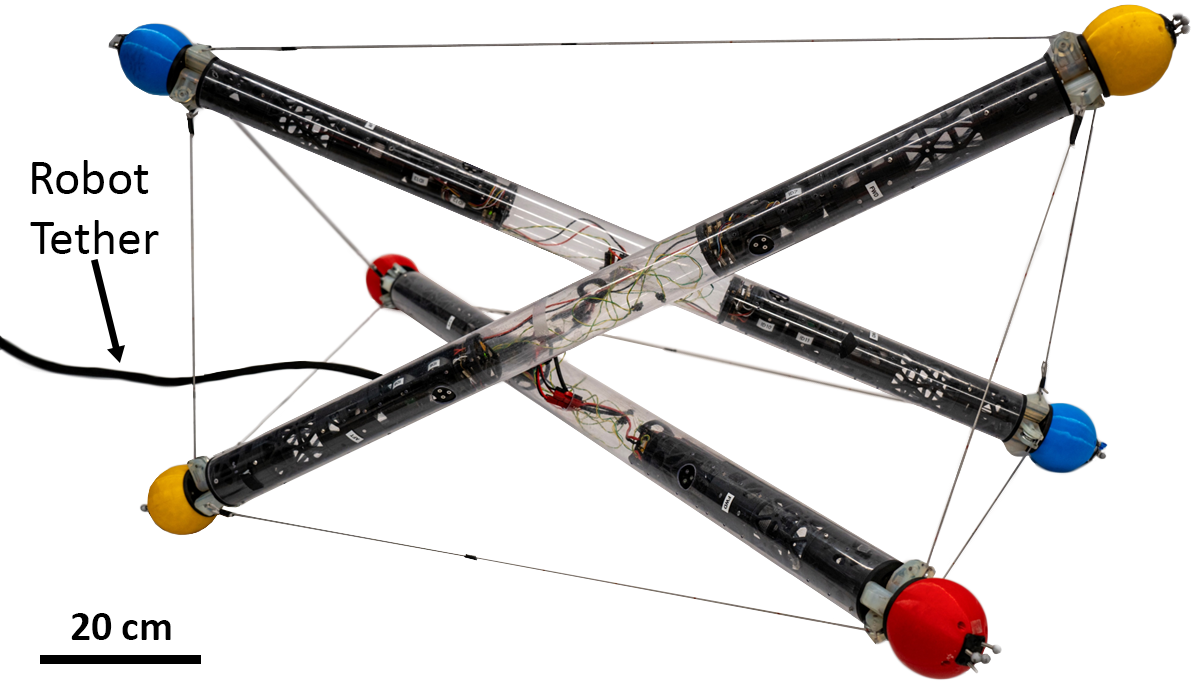}
    \caption{The assembled three-bar tensegrity robot. The robot tether enters through the triangular face of the structure and connects to the middle of each bar.}
    \label{fig:iso_real}
\end{figure}

The tensegrity robot proposed in this paper, shown in Fig. \ref{fig:iso_real}, advances towards full autonomy by addressing challenges in proprioception and bar space constraints. The design features on-the-fly stiffness tuning and precise control of cable length and force, achieved through the novel implementation of Quasi-Direct Drive (QDD) cable actuators. Proprioception is accomplished without relying on force sensors, torque sensors, or series elastic elements. The modular exoskeleton design provides a robust platform for integrating future sensing and on-board computing systems.

\begin{figure*}[ht]
    \centering
    \includegraphics[width=0.85\linewidth]{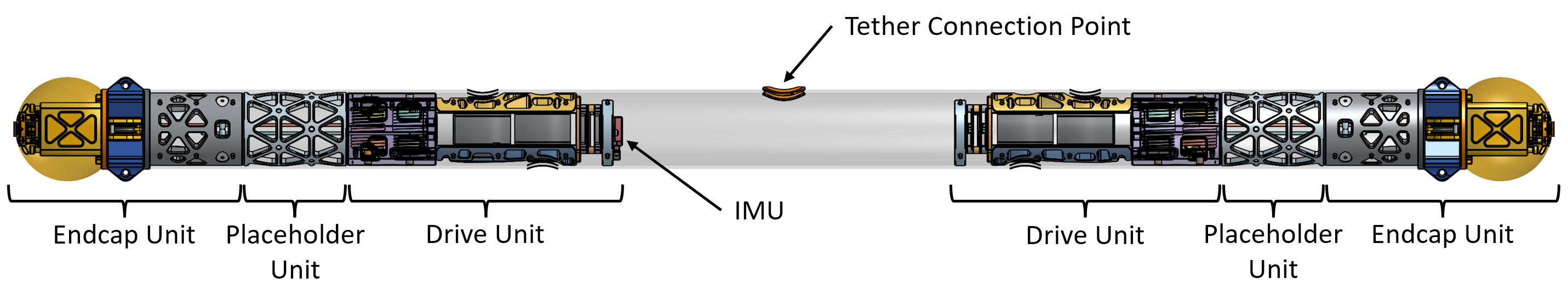}
    \caption{Internal view of a single tensegrity robot bar. Each bar contains two endcaps and two drive units. The inertial measurement unit (IMU) is mounted on the bottom of one of the drive units and has its z-axis aligned with the central axis of the bar. The placeholder unit inserted between the endcap and the drive unit demonstrates a feasible location for an additional module. Additional modules can also be inserted in the space between the two drive units.}
    \label{fig:barSideProfile}
\end{figure*}

The remainder of this paper is organized as follows: Section II provides a review of related works. Sections III, IV, and V detail the mechanical, electrical, and software design of the robot, respectively. Section VI presents the results from cable actuator and robot testing. Finally, Section VII summarizes the contributions of the paper and outlines the directions for future work.

\section{Related Works}
Due to the unconventional structure of tensegrity robots, a diverse range of different actuation strategies have been explored. 
Series elastic actuators (SEAs) \cite{prattSeriesElasticActuators1995} are widely used for cable actuation to provide structural compliance. Various approaches have been employed, including cable-spring assemblies \cite{sabelhausHardwareDesignTesting2014}, \cite{ceraEnergyEfficientLocomotionStrategies2019}, \cite{chenSoftSphericalTensegrity2017} and compliant cables \cite{paulDesignControlTensegrity2006a}, \cite{vespignaniDesignSUPERballV22018}.
However, these methods pose challenges for state estimation due to the difficulty of accurately measuring the length of elastic elements. Moreover, cable-spring assemblies either require an external spring, which can become tangled \cite{sabelhausHardwareDesignTesting2014}, or a large internal spring, which occupies a significant amount of space within the bar.
Elastomer-based tendons with capacitance-based sensing \cite{johnsonSensorTendonsSoft2022} offer accurate elastic length estimation when combined with cable actuators. However, they are challenging to manufacture and difficult to scale for larger robots. Additionally, they are susceptible to drift, creep under constant stress, and damage from sharp objects or obstacles.

Several works have achieved shape changing by varying bar length with pneumatic and linear actuators \cite{kimRobustLearningTensegrity2015}, \cite{kimRapidPrototypingDesign2014}, \cite{kimRobustLearningTensegrity2015}, \cite{yagiEvaluationShapeChangingTensegrity2019}. However, linear actuators are not ideal bar replacements due to their susceptibility to damage from shock loads. Besides, this method of locomotion generally results in slower speeds compared to other techniques. For example, the TT2 robot proposed by Kim \textit{et al.} achieved a maximum speed of only 0.014 body lengths per second (BLPS), or 0.01 m/s \cite{kimRapidPrototypingDesign2014}. Pneumatic skins \cite{bainesRollingSoftMembraneDriven2020} have proven effective but, like elastomer-based tendons, are difficult to scale up and require large off-board compressors. 

Vibration actuators offer a promising alternative for fast and low-complexity locomotion compared to traditional bar or cable actuators \cite{khazanovExploitingDynamicalComplexity2013}. Rieffel \textit{et al.} demonstrated speeds of up to 0.88 BLPS (0.115 m/s). 
However, intense vibrations are problematic for computer vision systems, which struggle with high-frequency perturbations \cite{sayedImprovedHandlingMotion2021}, \cite{xuPerfectNoisyWorld2024}. Furthermore, vibration-based locomotion cannot handle obstacle traversal. Consequently, this approach is incompatible with the goals of developing autonomous and intelligent tensegrity robots.

Our work advances the cable and spool actuated architecture by introducing QDD actuators, which are well-established in modern legged and humanoid robots. These actuators are favored in these fields due to their ability to control joint stiffness and sense external forces without relying on external torque sensors, force sensors, or series elastic elements \cite{wensingProprioceptiveActuatorDesign2017}.





\section{Mechanical Design}

While the tensegrity robot depicted in Fig. \ref{fig:iso_real} features a three-bar configuration, this design is not limited to just three bars. Each tensegrity bar is a standalone unit, allowing the system to be easily extended to configurations with four, six, or twelve bars.
The bar exoskeleton is constructed from a polycarbonate tube with a length of 1.2 m, a diameter of 76 mm, and a wall thickness of 3 mm. Polycarbonate is selected for its excellent impact resistance and ease of fabrication. Additionally, its transparency facilitates easier debugging by providing clear visibility of the internal components. 
This tube exoskeleton serves as the base upon which different combinations of modules can be inserted to achieve the desired robotic functionalities.

Currently, two fundamental modules have been developed: the dual drive unit and the endcap unit, both required for basic robot locomotion. An assembled bar, as illustrated in Fig. \ref{fig:barSideProfile}, incorporates two of each of these units. With each bar weighing 4.0 kg, the total weight of the three-bar robot is 12.0 kg. The mechanical design of the robot is detailed in the following.

\subsection{Dual Drive Unit}
The dual drive unit houses the powertrain responsible for actuating the cables of the robot. Each tube accommodates two such units, enabling a maximum of up to four actuated cables per bar. 

\begin{figure}[t]
    \centering
    \includegraphics[width=0.85\linewidth]{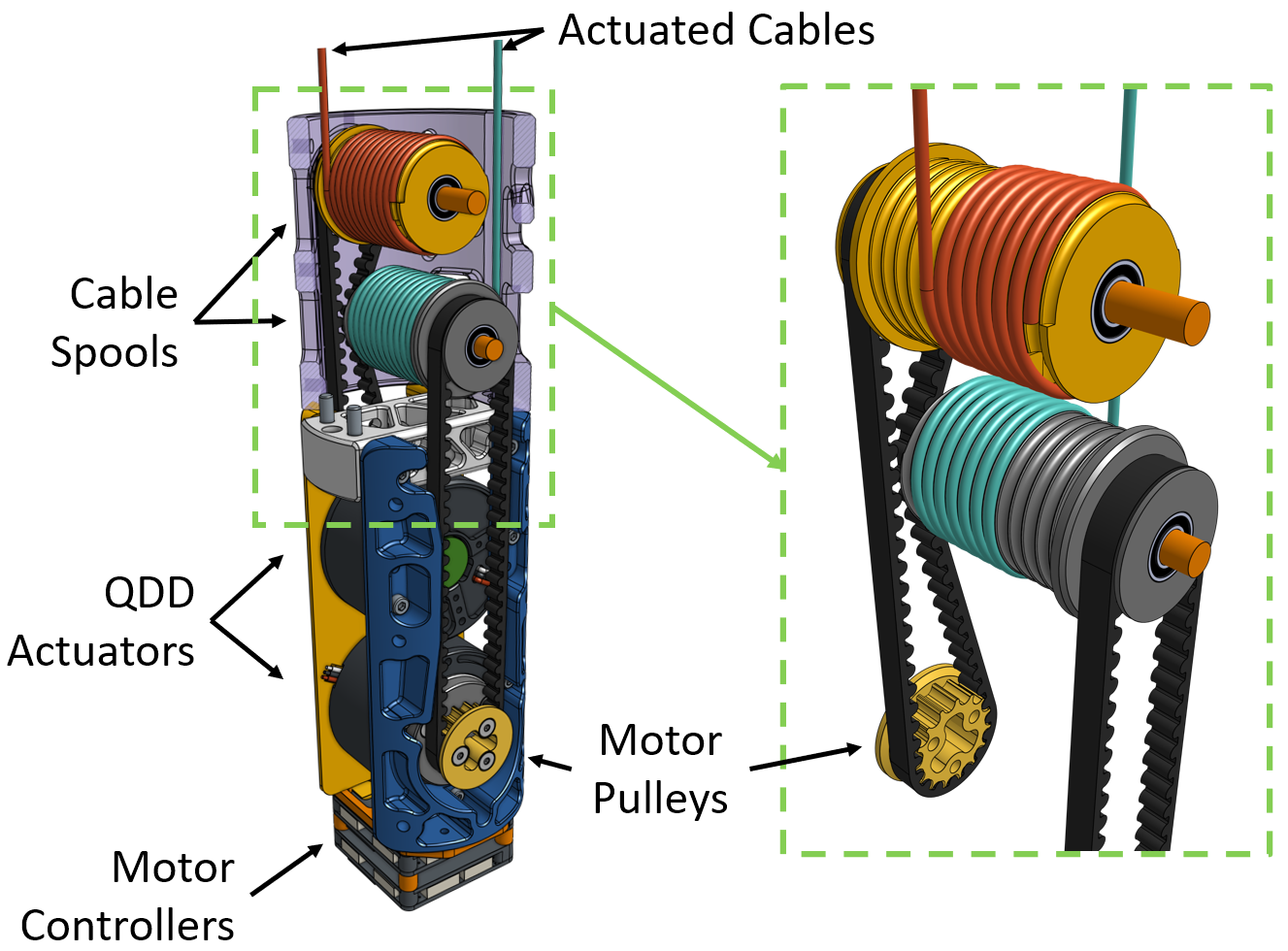}
    \caption{Cutaway view of the dual drive unit model. The red and teal strings are the actuated cables. The right side of the image shows a close up of the belt driven, dead-axle, grooved cable spool design. The motor controllers are located at the base of the unit.}
    \label{fig:DDU}
\end{figure}

To achieve variable cable length, tension, and operation mode while maintaining precise cable length sensing, QDD actuators are employed. QDD actuators are required for high-speed force control \cite{seokActuatorDesignHigh2012} of the cables. They enable the robot to replicate the operation of an SEA without the cable length inaccuracies associated with SEAs when using purely encoder-based cable length feedback. The design, illustrated in Fig. \ref{fig:DDU}, does not require additional force or torque sensors.

\subsubsection{Actuator Design}

\begin{figure}[t]
    \centering
    \includegraphics[width=0.4\linewidth]{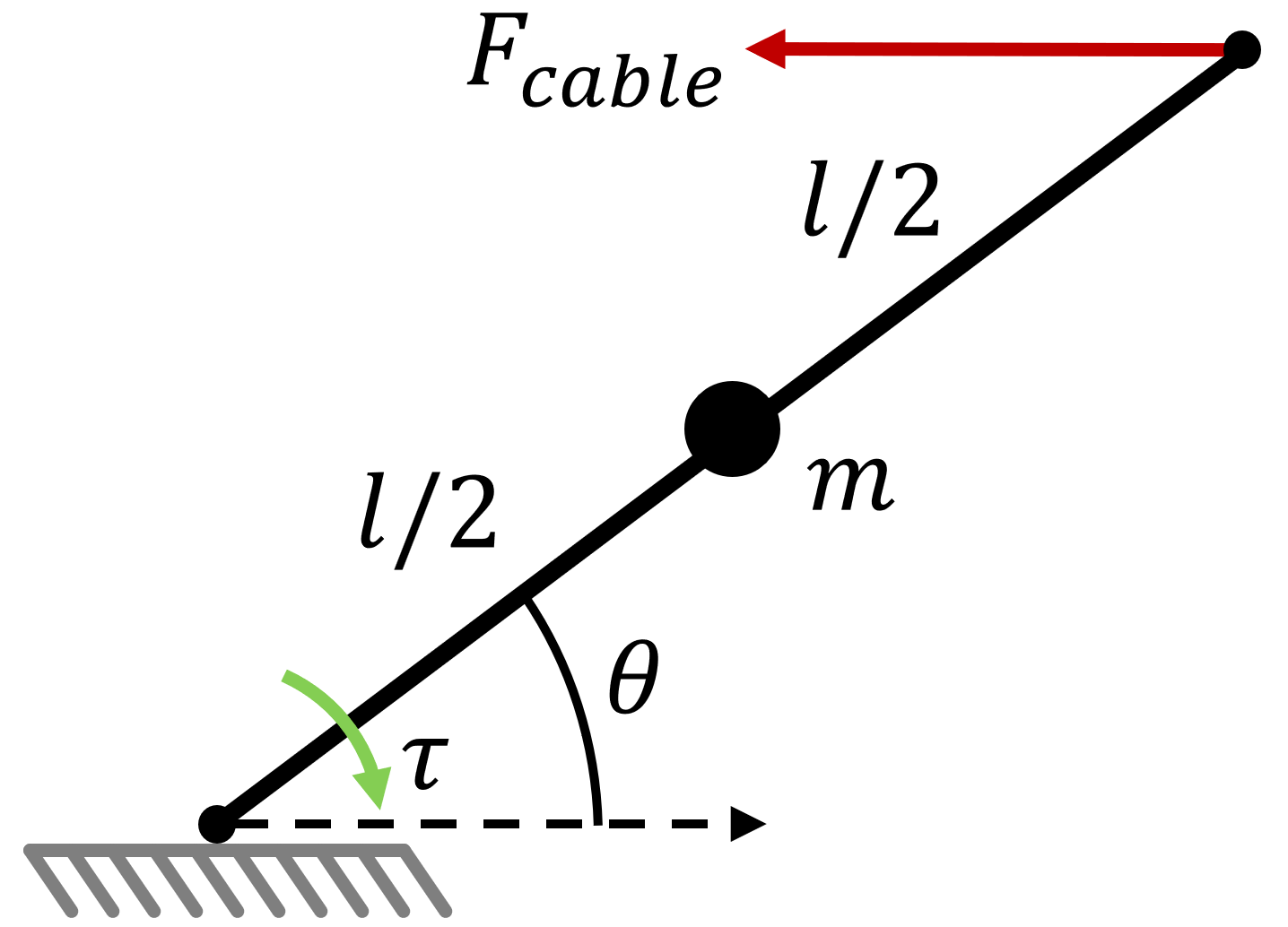}
    \caption{Simplified 2D physics model for a single bar held in equilibrium by a cable and ground contact.}
    \label{fig:2DPhysics}
\end{figure}

The expected cable forces are modeled to determine the actuator torque requirements. A simplified 2D model of the robot, shown in Fig. \ref{fig:2DPhysics}, is used to approximate the maximum required cable force for the robot to support itself. 
In this simplified model, a single bar pivoting around a fixed point is considered. The bar is treated as a point mass since, in a future untethered version, the center of the bar will house a battery, which will constitute a significant portion of the bar's mass. The cable force is on the floating end and pulls horizontally on the bar to keep it in equilibrium. 
The torque on the bar, $\tau_b$, can be estimated with:

\begin{equation}
    \label{eq:2dBarTorque}
    \tau_b = F_g\frac{l_b}{2}\cos(\theta)
\end{equation}
where $F_g$ is the weight of the bar, $\theta$ is the angle formed by the bar and the ground, and $l_b$ is the length of the bar. The required cable force, $F_c$, to keep the bar in equilibrium is then:

\begin{equation}
    \label{eq:2dCableForce}
    F_c = \frac{\tau_b}{l_b}\frac{1}{\sin(\theta)}
\end{equation}

Substituting (\ref{eq:2dBarTorque}) into (\ref{eq:2dCableForce}) results in the required cable force as a function of bar weight and angle:

\begin{equation}
    \label{eq:2dCableForceFinal}
    F_c = \frac{F_g}{2}\frac{\cos(\theta)}{sin(\theta)}
\end{equation}

The maximum force exerted on the cable is constrained by the most horizontal angle of the bar, which is restricted by bar-to-bar interference that can occur when the robot is in its flattest position. Based on the geometry of the robot, we estimate this angle to be 8 degrees. With an estimated bar weight of 5 kg, the maximum force required on the cable to support the bar is 174 N. The torque on the motor can be related to the force on the cable with: 

\begin{equation}
    \label{eq:torqueToForce}
    F_c = \frac{\tau_m}{r_s}
\end{equation}
where $r_s$ is the radius of the spool and $\tau_m$ is the torque on the output of the motor gearbox. The cable spool has a radius of 0.015 m, meaning the actuator must be able to provide a continuous torque of at least 2.61 Nm to self-support at the flattest robot position.

The chosen actuator is the GIM4310 brushless direct current (BLDC) motor, which has an integrated 10:1 planetary reduction gearbox \cite{gim4310}. It provides a continuous torque of 3 Nm and a stall torque of 6 Nm, meeting the torque requirements. The single-stage planetary gear reduction is the only reduction in the system, contributing to a relatively transparent transmission---an important consideration in the design of QDD actuators.

\subsubsection{Cable Spool}
The robot's actuated cables are attached to custom 30 mm diameter grooved spools with integrated pulleys. The grooved spool allows for more controlled winding compared to a bare spool, which improves cable length measurement accuracy \cite{guenersDesignImplementationCabledriven2022}. Each spool can wind and unwind 1.2 m of 2 mm diameter cable.
The integrated pulley and dead-axle design of the spool removes backlash that would otherwise be present in a live-axle design. The end of the cable is held in place with a knot in the spool. 
A close up of the cable spool design is shown in Fig. \ref{fig:DDU}.

\subsubsection{Cable Selection}


Various cable materials have been employed in tensegrity robots and cable-driven parallel robots, both of which utilize cable spool actuators for motion and force transmission.
Although steel cable has traditionally been used in cable-driven robots, modern polymers such as ultra-high-molecular-weight polyethylene (UHMWPE) have proven to be both stronger and lighter than steel cables of the same diameter and length \cite{mckenna2004handbook}.
For meter-scale robots, the choice of cable material is particularly critical, as compared to centimeter-scale systems, due to the significantly larger forces involved in locomotion and impacts.
The strength, weight, and elongation of several cable options are presented in Table \ref{tab:cableTable}. Cable elongation is given at 30\% of the cable's maximum breaking load (MBL).

\begin{table}[h]
    \centering
    \caption{Comparison of Cable Materials}
    \begin{tabularx}{\columnwidth}{@{} l X X X l @{}}
        \toprule
        Material & MBL (kN) & Weight (g/m) & Elongation (\%) & Ref. \\ 
        \midrule
        \diameter 2 mm Steel & 2.75 & 15.2 & $<$ 2 & \cite{sabelhausHardwareDesignTesting2014} \\
        \diameter 4 mm Nylon & 3.4 & 9.02 & 13-15 & \cite{vespignaniDesignSUPERballV22018}\\
        \diameter 1.4 mm Vectran & 2.2 & 6.8 & $<$ 1 & \cite{bruceDesignEvolutionModular2014}\\
        \diameter 2 mm Dyneema & 4.5 & 3.72 & $<$0.7 & This work\\
        \bottomrule
    \end{tabularx}
    \label{tab:cableTable}
\end{table}

For the tensegrity robot presented in this paper, HTS-75 12-strand Dyneema\textregistered{} SK-78 UHMWPE cord with a diameter of 2 mm is selected. The Dyneema cord offers a tensile strength of 4.5 kN and less than 0.7\% elongation at 30\% of its MBL.
Compared to steel cable, the flexibility of the polymer cable allows for the use of smaller pulleys and cable spools. For optimal performance, steel cables require pulleys that are 25 times larger than the cable diameter \cite{pott2013cable}, while polymer cables only need pulleys that are 10 times the cable diameter \cite{guenersCableBehaviorInfluence2021a}. Therefore, without using polymer cable, it would have been impossible to package all the required components into the exoskeleton tube.
When compared to the nylon cord used in SUPERballv2 \cite{vespignaniDesignSUPERballV22018}, the elongation factor of the Dyneema cord is significantly smaller, leading to better cable length estimation. 
Vectran is another excellent choice of polymer cable due to its low stretch properties, but it is difficult to acquire and cost-prohibitive.

\subsection{Endcap Unit}
The endcap unit is the contact point between the robot and the environment and is responsible for routing cables from the drive units to the anchor points on other bars. Ensuring smooth cable routing is important for minimizing cable friction to achieve more accurate cable length estimation.



\subsubsection{Compliant Cap}

\begin{figure}[t]
    \centering
    \includegraphics[width=0.34\linewidth]{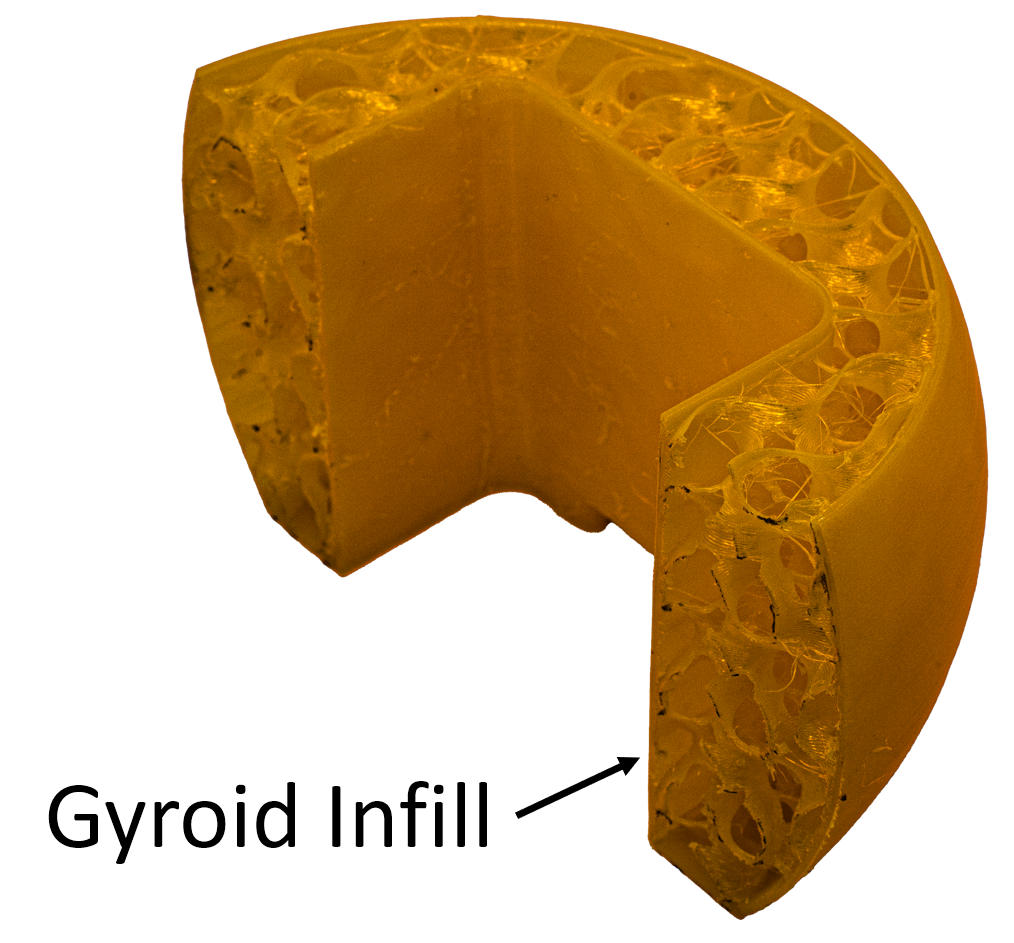}
    \caption{A sliced open compliant cap shows the internal structure. The low density gyroid infill allows for uniform compliance in all directions. The part is printed with 95A TPU filament.}
    \label{fig:endcap1}
\end{figure}

The sphere on the endcap is 3D printed using 95A thermoplastic polyurethane (TPU) with a 7\% gyroid infill pattern. This low infill percentage combined with the gyroid pattern ensures the sphere is uniformly compliant in all directions, providing adequate cushioning and grip. A sectional view of the printed cap is shown in Fig. \ref{fig:endcap1}. The center of the cap has a rigid internal frame to support future sensor attachment, such as contact sensors, force sensors, or even cameras.

\subsubsection{Cable Router and Cable Anchor}

The cable routing system consists of swiveling cable pulleys and anchors designed to facilitate the smooth transfer of cables between bars. 
Each cable router utilizes a 20 mm diameter cable pulley to redirect cables from the cable spool to the next adjacent bar. 
The router swivels horizontally on a shoulder bolt, with a wide 128-degree range that ensures smooth cable movement across all robot shape configurations.
A stainless steel pin, positioned just above the pulley, prevents dislodging of the cable from the pulley groove.

\begin{figure}[t]
    \centering
    \includegraphics[width=0.85\linewidth]{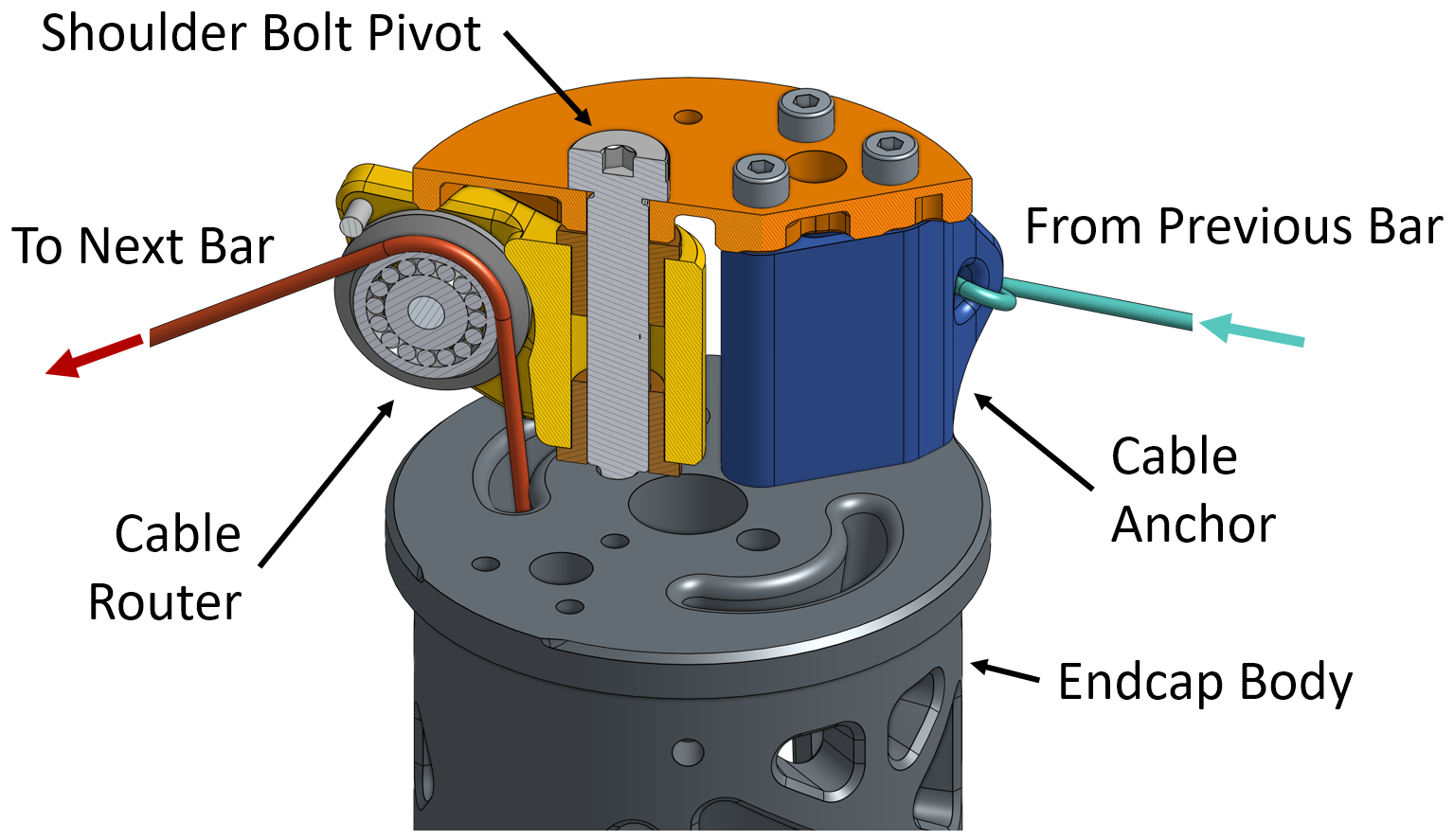}
    \caption{Sectional view of the endcap cable router and cable anchor model. One of the cable routers is on the left in yellow and one of the cable anchors is on the right in blue. The sectional view shows the swiveling and cable guide mechanism of the cable router. The compliant cap is mounted on top of the orange plate.}
    \label{fig:endcap2}
\end{figure}

The cable anchor serves dual purposes: it functions as the attachment point for the end of each actuated cable and acts as a spacer between the endcap body and the compliant cap mounting plate.
A sectional view of the cable router and cable anchor is shown in Fig. \ref{fig:endcap2}. Each endcap incorporates one pair of cable routers and one pair of cable anchors, arranged such that the routers and anchors are alternated and clocked 90 degrees apart from each other.

\begin{figure*}[ht]
    \begin{center}  \includegraphics[width=0.77\linewidth]{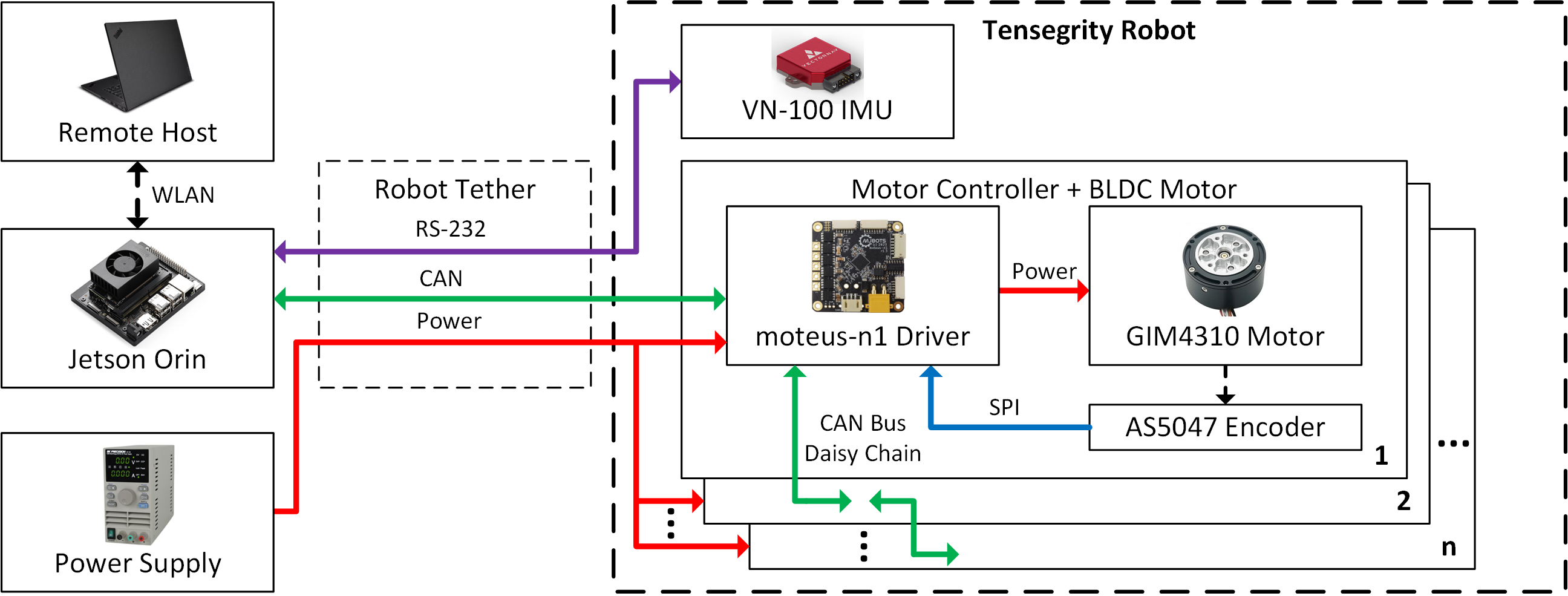} 
    \caption{Block diagram of the tensegrity robot electrical system. Robot power and control signals are wired through the robot tether. The controller area network (CAN) bus is daisy-chained from the Jetson Orin to each of the moteus motor controllers.}
    \label{fig:EEBlock}
    \end{center}
\end{figure*}

\subsection{Manufacturing}
The robot hardware is produced using a combination of metal machining, Fused Deposition Modeling (FDM) and Stereolithography Apparation (SLA) 3D printing techniques. 
Shafts and axles are machined from high-strength 7075-T6 aluminum. 
The bodies of the dual drive units and endcap units are printed with Polylactic Acid (PLA) filament. 
High-strength components, including the motor pulley, endcap body, and motor mounts, are fabricated using carbon fiber reinforced Nylon 6 (PA6-CF) filament through FDM printing. 
Components experiencing stress in two orthogonal directions, such as the cable router pulley brackets, cable anchors, and cable spools, are printed using Formlabs Durable resin \cite{durableResin} with SLA technology. 
This choice is due to SLA’s isotropic properties, which are better suited for handling multi-directional stresses compared to the anisotropic properties of FDM-printed components. 
Custom 3D-printed drilling jigs are used to drill the holes in the exoskeleton tube. 
FDM components are printed with a Bambu X1C printer, while SLA components are printed with a Formlabs Form 3+ printer.

\section{Electrical Design}

The electrical system, illustrated in Fig. \ref{fig:EEBlock}, includes a power supply, motor controllers, cable actuators, an inertial measurement unit (IMU) sensor, robot controller, host device, and cable harness. 
The following discusses the design of each component in detail.

\subsection{Robot Power}

The electronics are powered by a single 300 W bench power supply providing 24 V. Power is distributed through 18 AWG wires, with each bar receiving power from its own pair. These power wire are bundled with the signal wires within a sheathed wire harness, as illustrated in Fig. \ref{fig:iso_real}.


\subsection{Motor Controller}
MJBots moteus-n1 BLDC motor controllers \cite{moteus} are chosen to control the cable actuators. These controllers are selected for their ability to meet motor power requirements, small form factor, and controller area network (CAN) bus connectivity. The BLDC motors are operated using field oriented control (FOC) to enhance efficiency and controlability.
For torque feedback, low-side current sense resistors are used by the motor controller. The developed torque of the motor can be calculated using:  

\begin{equation}
    \label{eq:torqueCurrent}
    \tau_m = \frac{i_a}{K_v}n_r
\end{equation}
where $K_v$ is the motor's speed constant, which is obtained through a one-time self-calibration sequence, $i_a$ is the armature current. The motor controllers were modified to use 10 mOhm current sense resistors instead of the stock 0.5 mOhm resistors to improve current sensing and therefore torque sensing capabilities. 

\begin{figure*}[ht]
    \begin{center}  \includegraphics[width=0.82\linewidth]{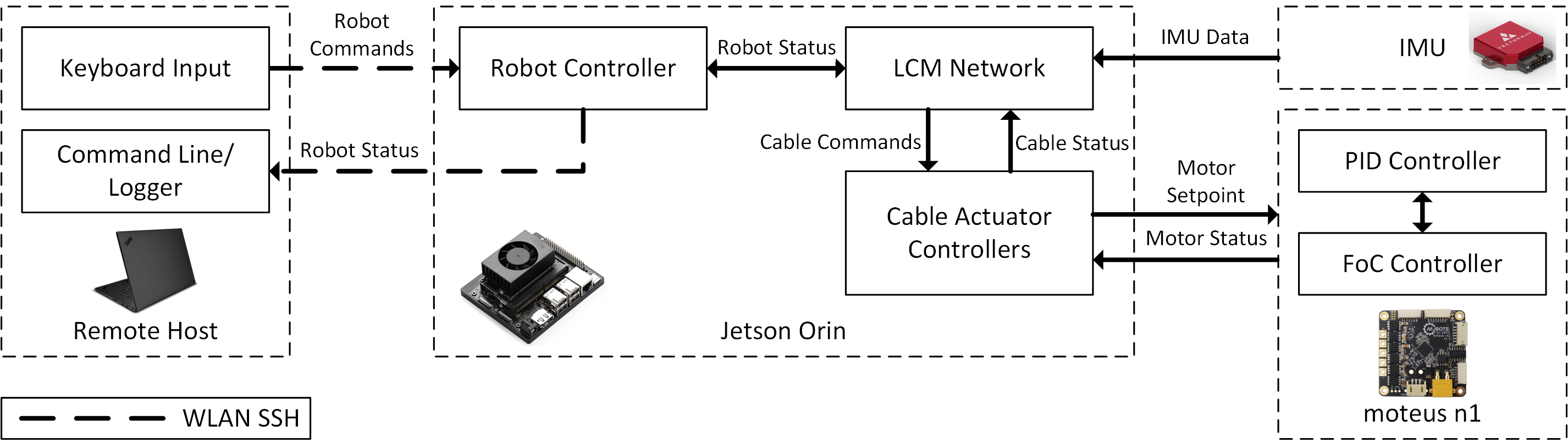} 
    \caption{Block diagram of the software stack. The remote host sends encoded control messages to the robot controller, which then communicates with the IMU and cable actuator controllers via the LCM network. The cable actuator controllers exchange motor setpoint and status data with the low-level motor controllers.}
    \label{fig:SWEBlock}
    \end{center}
\end{figure*}

For motor position feedback, an AS5047 absolute magnetic encoder connected to the motor controller through a 4-wire Serial Peripheral Interface (SPI) is utilized. The absolute position functionality is required for motor commutation. 
For cable length estimation, the encoder operates as a relative encoder and must be initialized to a home position when the robot is powered on. This initialization is achieved by manually setting each cable to the software specified initial length. The cable length $l_c$ can then be estimated using: 

\begin{equation}
    \label{eq:posToLength}
    l_c = 2 \pi r_s n
\end{equation}
where $n$ is the number of encoder turns. Due to space constraints, it is challenging to fit a multi-turn absolute encoder directly on the spool.

\subsection{Additional Sensors}

The integration of QDD actuators minimizes the number of sensors needed for proprioception. The sole additional sensor employed is a VectorNAV VN-100 inertial measurement unit (IMU), which communicates with the robot controller via RS-232. This communication cable is incorporated into the wire harness.

\subsection{Robot Controller}
The robot controller is a NVIDIA Jetson Orin, interfacing with the robot CAN bus via a Texas Instruments TCAN1042 CAN transceiver. Motor control signals are transmitted over the robot CAN bus, which is configured in a daisy-chain topology as shown in Fig. \ref{fig:EEBlock}. This topology extends from the Orin through each motor controller and terminates after the last motor controller.
The use of daisy-chain topology reduces wiring complexity, requiring only a single pair of CAN wires in the wire harness to connect all control signals to the robot controller.
To control the robot, a remote host is connected to the robot controller via a wireless local area network (WLAN). 

\section{Software Design}
The software stack to control the tensegrity robot consists of a high-level robot controller, cable actuator controllers, and low-level motor controllers. Fig. \ref{fig:SWEBlock} illustrates the block diagram of the software stack. Currently, basic teleoperation allows the robot to roll forward, backward, turn, and change stiffness.
\subsection{Cable Actuator Control}

The QDD cable actuator enables dynamic cable control that can be adjusted in real-time to suit different situations. For instance, if the IMU detects that the robot is falling, the robot can be made softer to better absorb the shock load. Conversely, if the robot is carrying a payload, it can be made stiffer to hold shape.
The following three cable actuator control modes have been developed:

\begin{enumerate}
  \item Length mode: The cable is set to a specified cable length $l_{sp}$ with a proportional derivative (PD) feedback loop that enforces a minimum cable force to ensure that the cable is always under tension:
    \begin{equation}
        \label{eq:lenMode}
        F_c = 
        \begin{cases} 
        k_p (l_c - l_{sp}) + k_d \frac{d(l_c - l_{sp})}{dt} + F_{f} & l_c \ge l_{sp} \\
        F_{f} & l_c < l_{sp}
        \end{cases}
    \end{equation}
    where $k_p$ and $k_d$ are the proportional and derivative gain values of the PD controller, and $F_{f}$ is the manually tuned feed forward force to account for frictional losses.
  \item Spring mode: The cable acts as a spring. The actuator will imitate the specified spring function:
    \begin{equation}
        \label{eq:sprMode}
        F_c = f(l_c)
    \end{equation}
    where $f(l_c)$ is any arbitrary function dependent on the current cable length. 
  \item Constant force mode: A constant force is applied to the cable, regardless of the current cable length.
\end{enumerate}


To achieve variable stiffness control, the control constants in both length mode and spring mode can be adjusted.
In length mode, $k_p$ and $k_d$ function as spring constants, allowing the stiffness of the system to be varied by altering these gain values. 
In spring mode, any arbitrary function can be employed to define cable force. For instance, a linear spring can be represented by the following: 

\begin{equation}
    \label{eq:linearMode}
    f(l_c) = k_{s}l_c
\end{equation}
where $k_{s}$ denotes the specified spring constant. Alternatively, a possible non-linear spring function can be represented as the following: 

\begin{equation}
    \label{eq:nonLinearMode}
    f(l_c) = k_1 l_c^{k_2} + k_3
\end{equation}
where $k_1$ is a scaling factor, $k_2$ is the exponential factor, and $k_3$ is the minimum cable force. These parameters can be dynamically adjusted to suit different operational scenarios.

\subsection{Teleoperation}
For teleoperation, the teleoperator uses the remote host to publish an encoded control message onto a Lightweight Communications and Marshalling (LCM) \cite{huangLCMLightweightCommunications2010} network. This message includes the timestamp, cable control parameters, and the cable actuator ID. The cable control parameters specify the cable control mode and associated variables, such as cable length and tension.
The robot controller then decodes the LCM messages, computes the motor setpoints, and sends commands to the moteus motor controllers over the CAN bus to achieve the calculated motor setpoints. The cable actuator controller then sends a status message back to the remote host for diagnosis and monitoring.

\section{Experimental Results}
Experiments were conducted to verify the cable length estimation accuracy and variable stiffness control. The full robot was also tested to verify locomotion and variable stiffness functionality.

\subsection{Cable Length Accuracy}

\begin{figure}[t]
    \centering
    \includegraphics[width=0.6\linewidth]{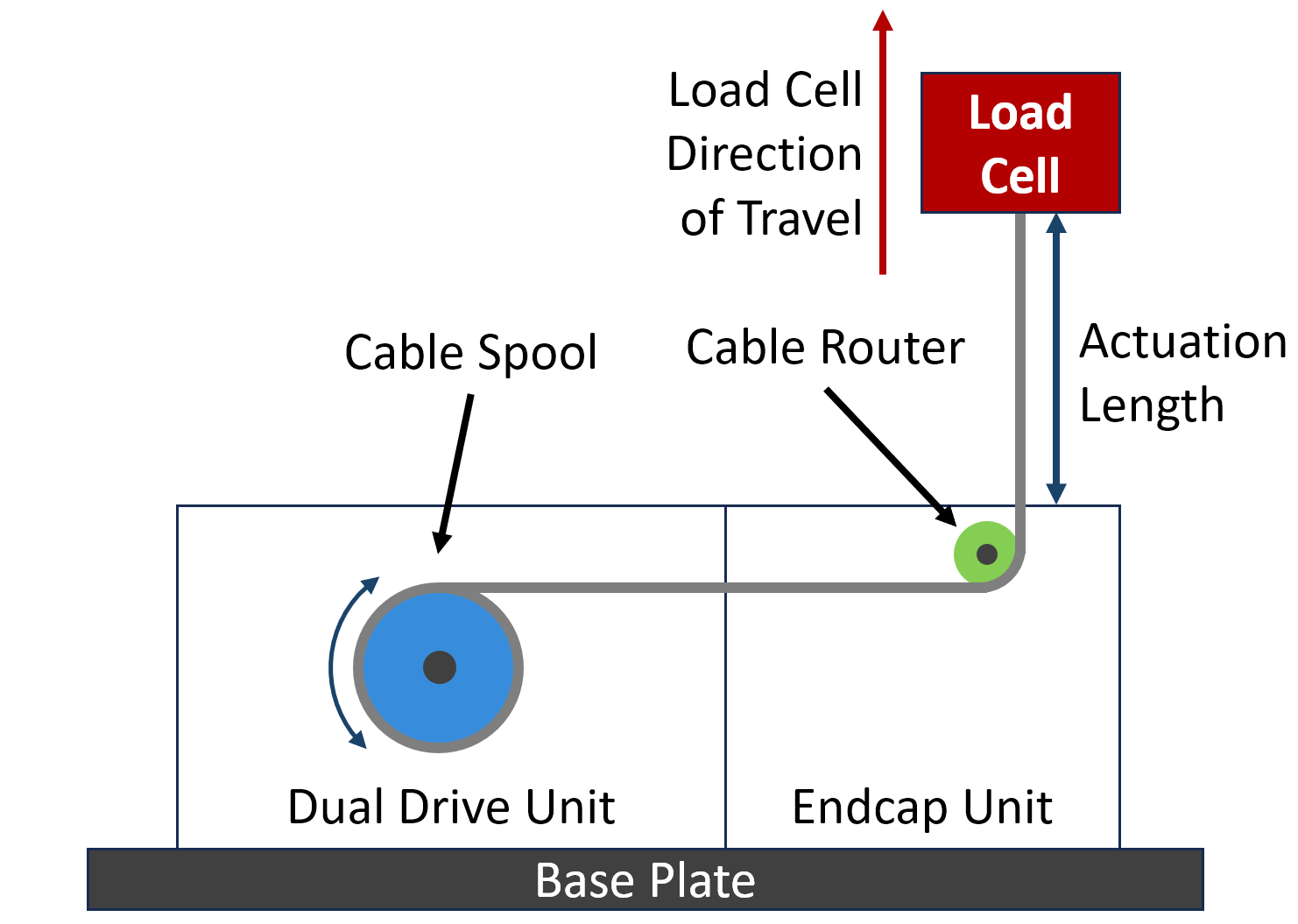}
    \caption{Cable length estimation and variable stiffness control experimental setup. A half-length tensegrity bar is attached to the base of the universal testing machine. The actuated cable is attached to the 5 kN load cell which can be moved up and down. The actuation length is defined as the distance from the load cell to the tip of the pulley bracket.}
    \label{fig:expSetup}
\end{figure}

Cable length estimation is critical for accurate state estimation. Errors in cable length estimation can arise from several sources, including cable stretch, cable wrap on the spool, and encoder resolution.
The experiment was conducted using the setup shown in Fig. \ref{fig:expSetup}. This setup comprised a half-length bar, which included one dual drive unit and one endcap unit. The half-bar was attached to the base of a universal testing machine (Instron 68SC-5), with the cable guide's output aligned to the center of the machine's load cell. The actuated cable was attached to the load cell using a carabiner clip.
To evaluate cable length estimation, the height of the load cell was adjusted from 0.05 m to 0.65 m in 0.05 m increments. The estimated cable lengths were recorded for loads of 50 N, 100 N, 150 N, and 200 N, which reflects the typical operational range of the cable actuator on the robot. Nylon paracord, used in Superballv2 \cite{vespignaniDesignSUPERballV22018}, was also tested for comparison.

\begin{figure}[t]
    \centering
    \includegraphics[width=0.55\linewidth]{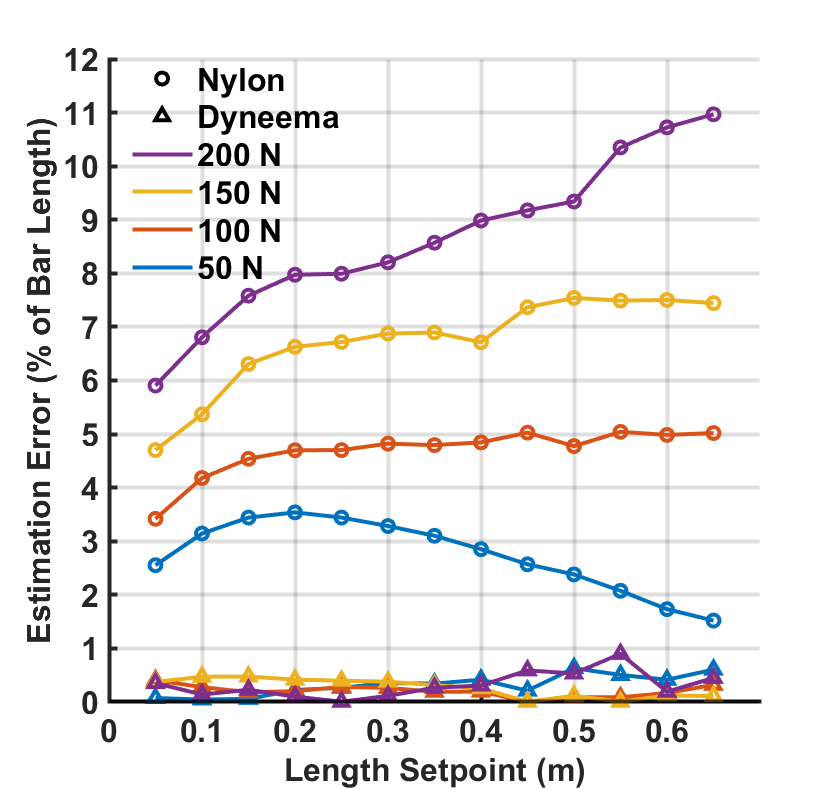}
    \caption{Cable length estimation error scaled as a percentage of bar length (1.22 m) over a range of different cable lengths and cable forces.}
    \label{fig:lengthError}
\end{figure}

For state estimation, it is more informative to express cable length error as a percentage of the bar length rather than as a percentage of the set point length. 
Results shown in Fig. \ref{fig:lengthError} indicate that the cable length estimation error with the Dyneema cable was less than 1\% of the robot bar length across the tested cable length and force range. 
On the other hand, the Nylon cable saw as much as 11\% error. The error also increased with force, which is the expected outcome of a more elastic cable.

Compared to some existing state estimation techniques, like external ranging sensors \cite{caluwaertsStateEstimationTensegrity2016b}, external cameras \cite{lu20226ndofposetrackingtensegrity}, or robotic skins \cite{boothSurfaceActuationSensing2021}, the proposed approach to cable length estimation demonstrates significantly improved accuracy without the need for additional sensors.

\begin{figure}[t]
    \centering
    \subfloat[]{
        \includegraphics[width=0.5\linewidth]{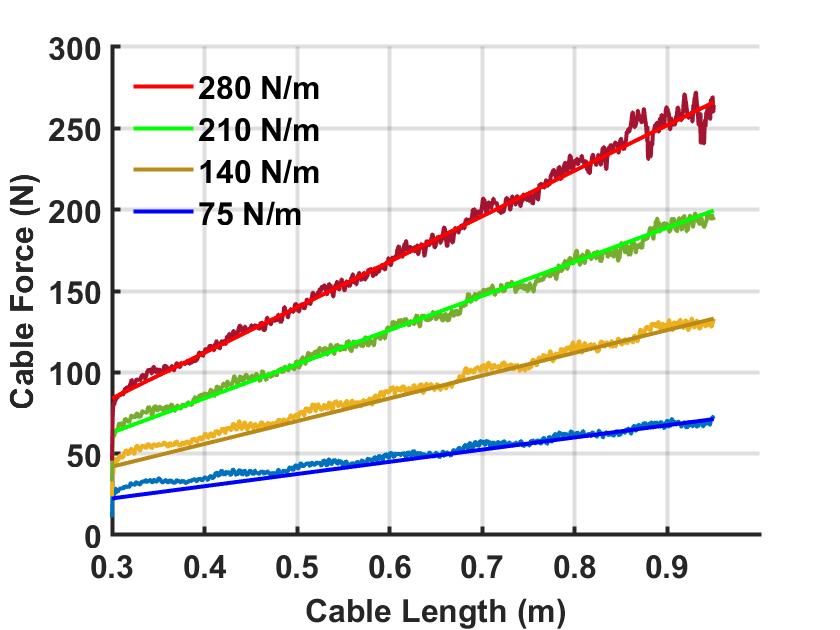}
        \label{fig:linearStiff}
        }
    \subfloat[]{
        \includegraphics[width=0.5\linewidth]{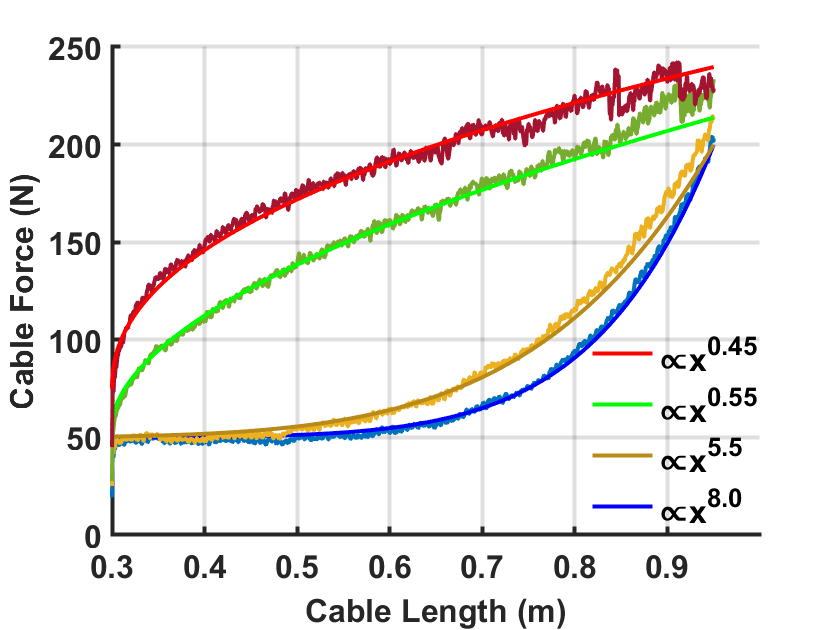}
        \label{fig:nonLinearstiff}
        }
    \caption{Variable stiffness control of cable force. (a) linear and (b) non-linear stiffness.}
    \label{fig:stiffControl}
\end{figure}

\begin{figure}[t]
    \centering
    \includegraphics[width=1\linewidth]{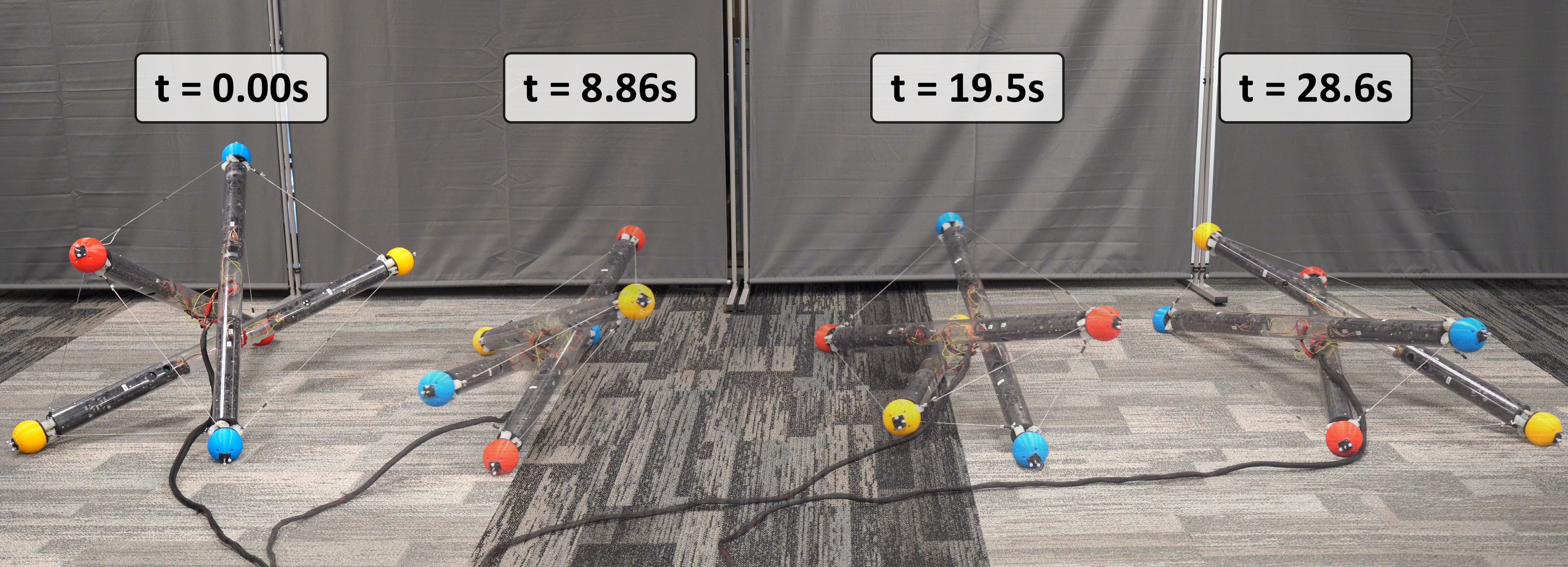}
    \caption{Motion sequence of the tensegrity robot rolling forward, showing distinct positions over time.}
    \label{fig:robotLocomotion}
\end{figure}

\begin{figure}[t]
    \centering
    \subfloat[]{
        \includegraphics[width=0.45\linewidth]{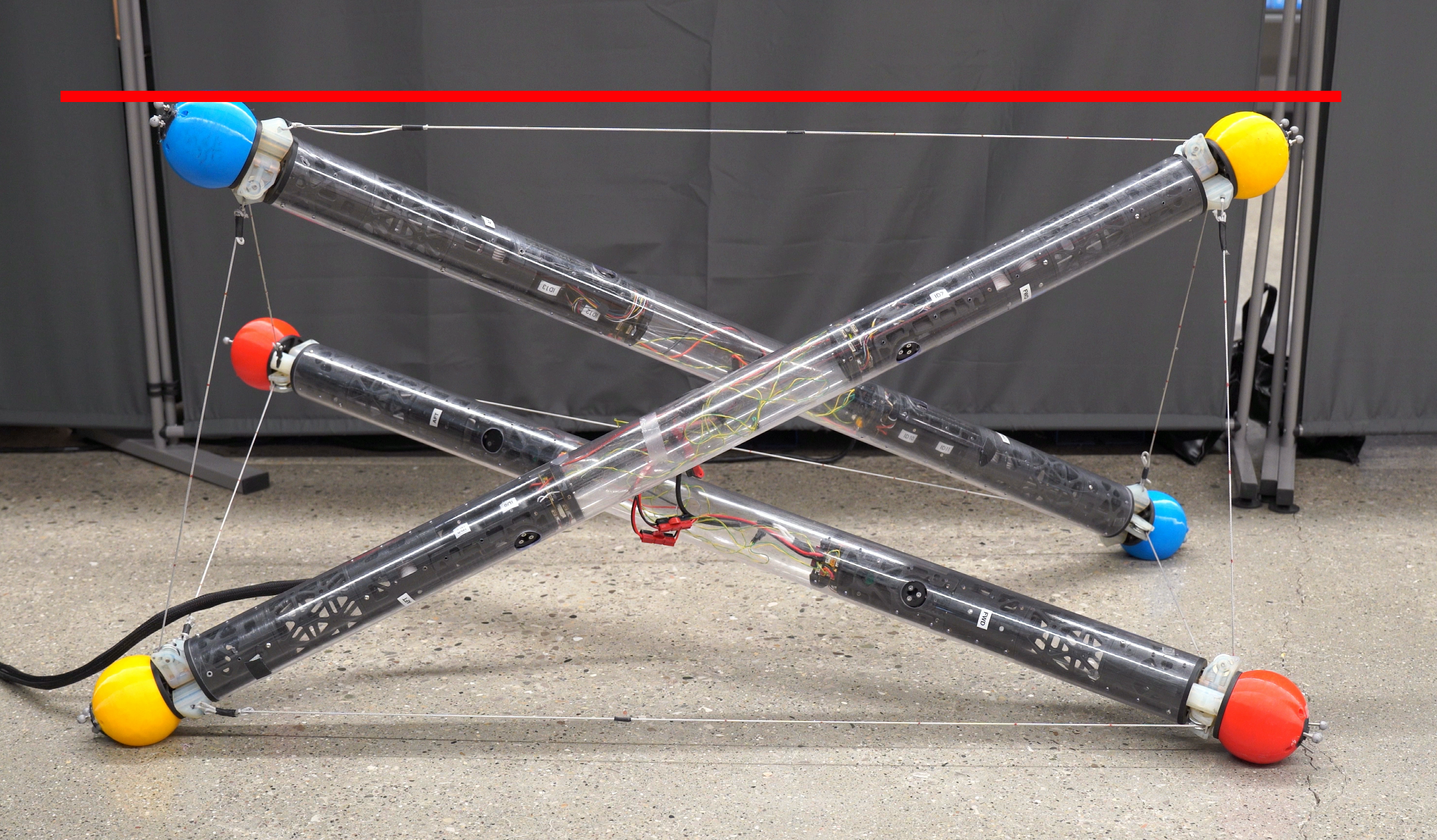}
        \label{fig:stiff1}
        }
    \subfloat[]{
        \includegraphics[width=0.45\linewidth]{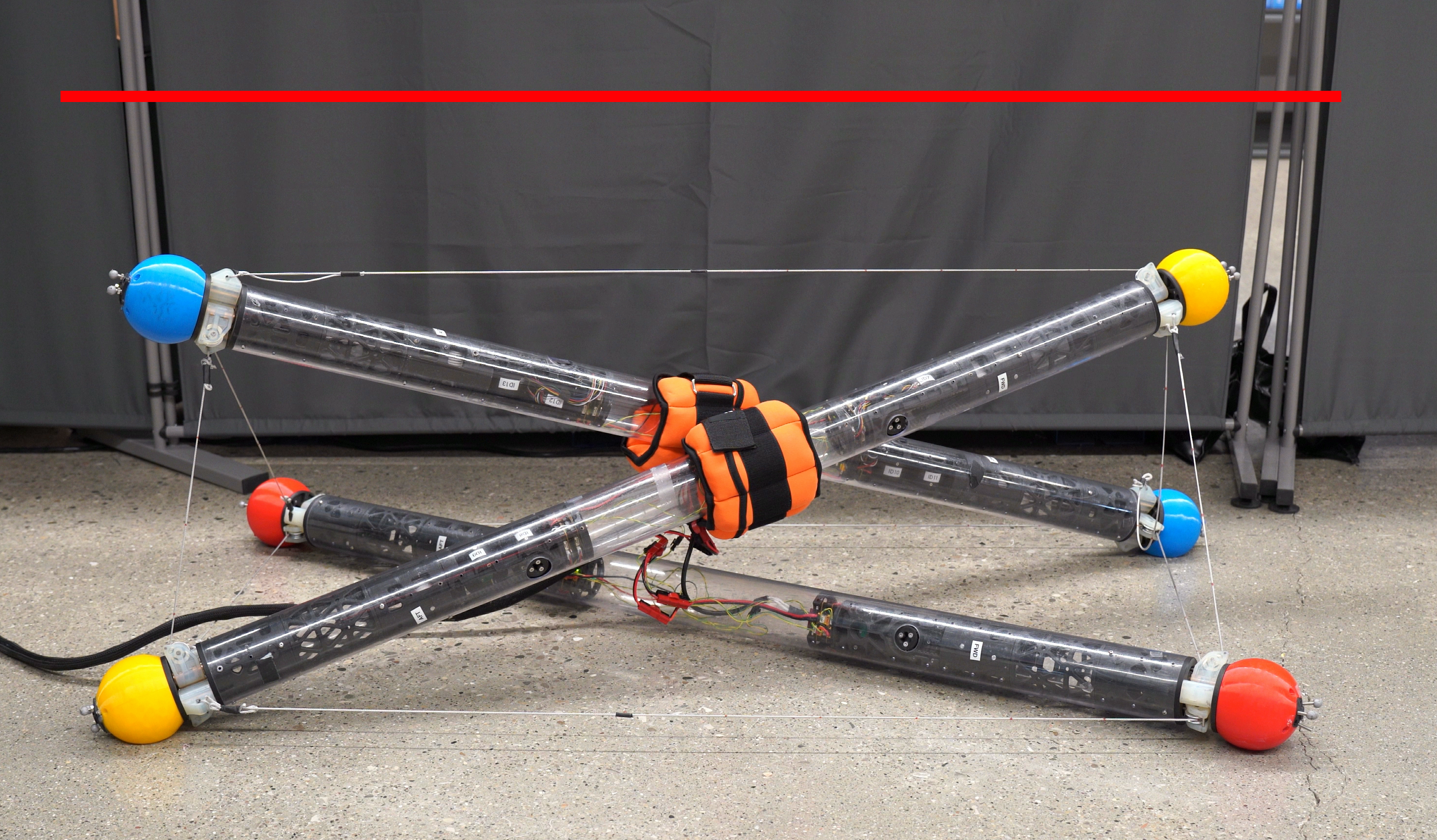}
        \label{fig:stiff2}
        }
    \qquad
    \subfloat[]{
        \includegraphics[width=0.45\linewidth]{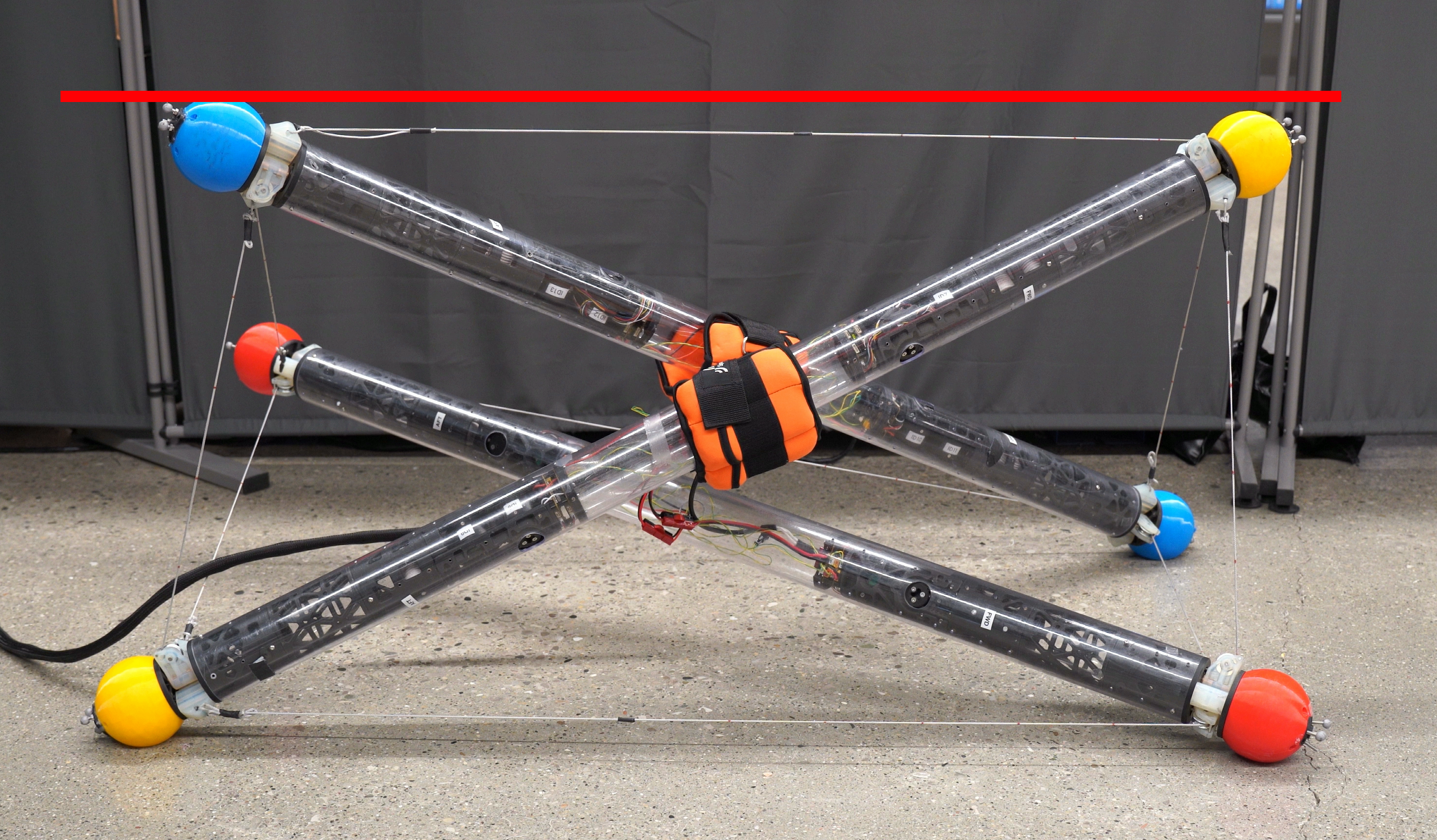}
        \label{fig:stiff4}
        }
    \subfloat[]{
        \includegraphics[width=0.45\linewidth]{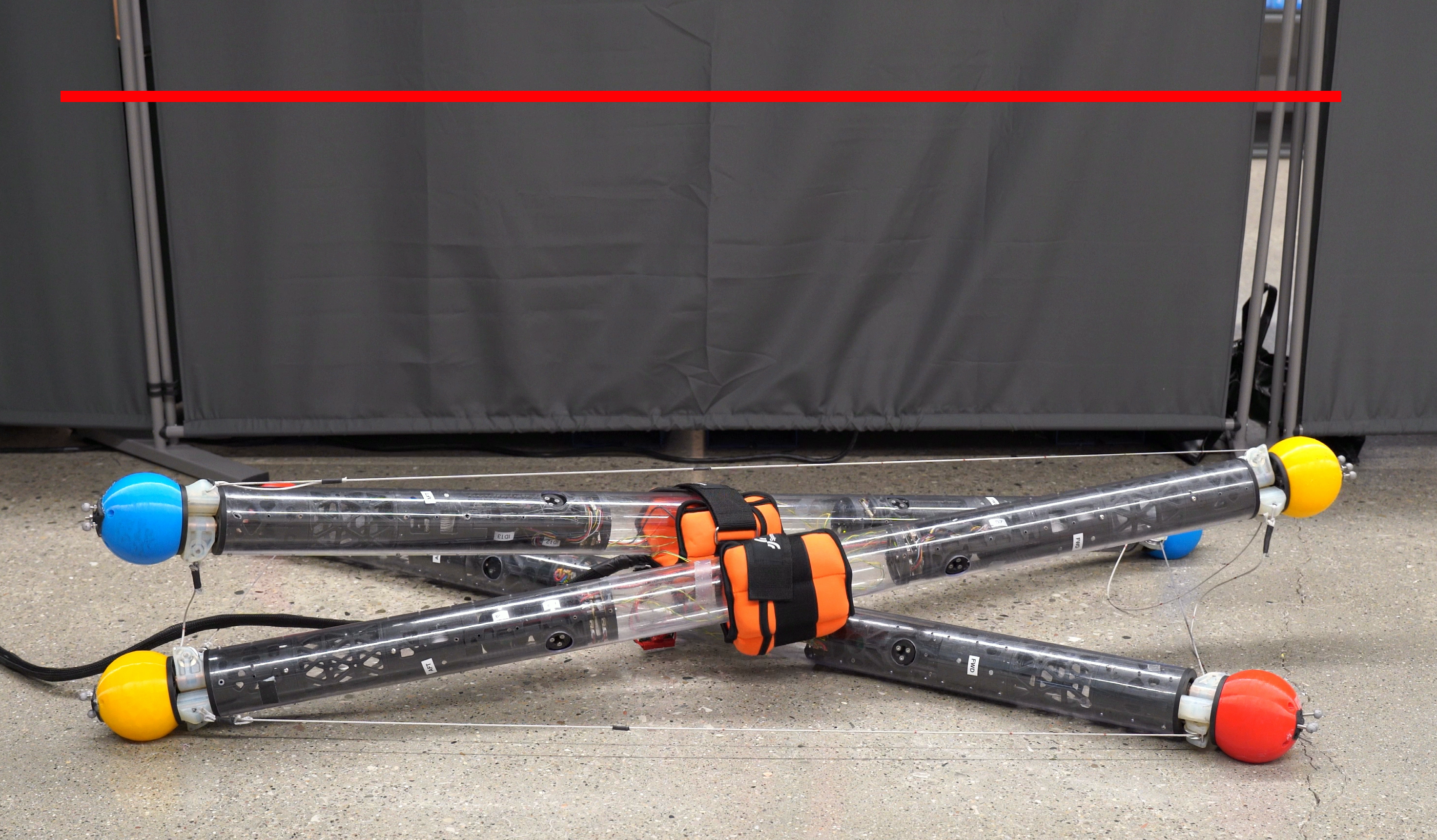}
        \label{fig:stiff3}
        }
    \caption{Variable stiffness control of the robot with constant target shape. The target shape's height denoted by the horizontal red line. The robot starts at 200 N/m stiffness with (a) no payload, and (b) 9 kg payload (orange). The stiffness is then increased to (c) 450 N/m to return to starting shape, and then decreased to (d) 100 N/m.}
    \label{fig:stifftest}
\end{figure}

\subsection{Variable Stiffness Control}

Variable stiffness control of the cable actuator was evaluated using the same experimental setup as the cable length accuracy test. 
The force response of the cable actuator was assessed by pulling the cable at a rate of 200 mm/s. A range of spring constants from 75 N/m to 280 N/m was tested, along with four different non-linear spring stiffnesses with varying exponential factors. For the linear stiffness, shown in Fig. \ref{fig:linearStiff}, the root mean squared error (RMSE) values were 3.75 N, 3.89 N, 3.28 N, 4.63 N for 75 N, 140 N, 210 N, and 280 N stiffness, respectively. For the non-linear stiffness, shown in Fig. \ref{fig:nonLinearstiff}, the 
 RMSE values were  2.81 N, 4.68 N, 5.66 N, and 4.73 N for $x^{8.0}$, $x^{5.5}$, $x^{0.55}$, $x^{0.45}$ stiffness curves, respectively. 
These results demonstrate that the cable actuator can be configured to behave as either a linear or non-linear spring.

\subsection{Robot Testing}

Robot locomotion was tested to verify the functionalities of the cable drive actuator (SI Video S1).
The motion sequence illustrated in Fig. \ref{fig:robotLocomotion} shows the different positions of the robot during rolling locomotion. 
The robot was tele-operated at 20Hz and rolls 1.67 revolutions to traverse 5.2 m in 28.6 s (0.15 BLPS). 
Additional testing showed the robot capable of locomotion with a heavy (11 kg) payload. (SI Video S2).

On-the-fly stiffness tuning was tested by observing the robot shape for varying stiffness and payload (SI Video S2). 
In Fig. \ref{fig:stiff1}, the robot begins at 200 N/m cross cable stiffness with no payload, allowing it to achieve the target shape. 
Then, in Fig. \ref{fig:stiff2}, a 9 kg payload is attached, causing the robot to sag under the added weight.  
The stiffness is subsequently increased to 450 N/m in Fig. \ref{fig:stiff4}, enabling the robot to return to its target shape. 
Finally, in Fig. \ref{fig:stiff3} the stiffness is lowered to 100 N/m, causing the robot to sag significantly. 
This variable stiffness control allows the robot to adjust its compliance according to different payload weights.

\section{Conclusion and Future Works}
In this paper, we presented a novel design for a variable stiffness tensegrity robot utilizing QDD cable actuators. The QDD actuators significantly improve cable length estimation accuracy and enable on-the-fly stiffness tuning without the need for external torque or force sensors. 
Experimental results demonstrate the robot's ability to achieve $<$1\% cable length estimation error relative to bar length and force control of an arbitrary stiffness curve with $<$6 N of RMSE. The modular exoskeleton design provides a robust platform for integrating sensor and computing modules, opening a pathway for fully autonomous and intelligent tensegrity robots. 

Future work will focus on incorporating additional computing, sensing, and camera modules to enable untethered autonomous operation. We will also conduct robustness testing and present comprehensive state estimation results. 
New functionalities enabled by variable stiffness control will also be also be explored to optimize locomotion and impact resistance based on environment and payload conditions.
Developing an open-source standard for module design will help enable collaborative, custom modules tailored to mission specific requirements.


\section*{ACKNOWLEDGMENT}

The authors would like to acknowledge Xiaohao Xu, Jiaqi Wang, and Eeshwar Krishnan for their insightful feedback and help with robot fabrication throughout the design and experimental process.


\bibliographystyle{IEEEtran}
\bibliography{main.bib}

\end{document}